\documentclass[journal]{IEEEtran}
\ifCLASSINFOpdf
\else
\fi

\usepackage{graphicx}
\usepackage{url}
\usepackage{booktabs}
\usepackage{breakurl}
\usepackage{multirow}
\usepackage{array}
\usepackage{tabu}
\usepackage{xcolor}
\usepackage{amsmath}
\usepackage{amssymb}
\usepackage{comment}
\usepackage{algorithm}
\usepackage[noend]{algpseudocode}

\makeatletter
\def\algbackskip{\hskip-\ALG@thistlm}
\makeatother

\usepackage[hang,flushmargin]{footmisc}

\makeatletter
\def\footnoterule{\relax%
 \kern-5pt
 \hbox to \columnwidth{\hfill\vrule width \columnwidth height 0.6pt\hfill}
 \kern4.6pt}
\makeatother

\begin{document}
%
\title{Fingerprint Spoof Generalization}
%
%

\author{Tarang Chugh*,~\IEEEmembership{Student Member,~IEEE,} and~Anil K. Jain,~\IEEEmembership{Life~Fellow,~IEEE}%
        
\thanks{T. Chugh and A. K. Jain are with the Department of Computer Science and Engineering, Michigan State University, East Lansing, MI, 48824. E-mail: \{chughtar, jain\}@cse.msu.edu}%
\thanks{*Corresponding Author}
\thanks{A preliminary version of this paper was presented at the International Conference on Biometrics (ICB), Greece, June 4-7, 2019~\cite{gajawada2019universal}.}
}
\maketitle
\begin{abstract}
We present a style-transfer based wrapper, called Universal Material Generator (UMG), to improve the generalization performance of any fingerprint spoof detector against spoofs made from materials not seen during training. Specifically, we transfer the style (texture) characteristics between fingerprint images of known materials with the goal of synthesizing fingerprint images corresponding to unknown materials, that may occupy the space between the known materials in the deep feature space. Synthetic live fingerprint images are also added to the training dataset to force the CNN to learn generative-noise invariant features which discriminate between lives and spoofs. The proposed approach is shown to improve the generalization performance of a state-of-the-art spoof detector, namely Fingerprint Spoof Buster, from TDR of 75.24\% to 91.78\% @ FDR = 0.2\%. These results are based on a large-scale dataset of 5,743 live and 4,912 spoof images fabricated using 12 different materials. Additionally, the UMG wrapper is shown to improve the average cross-sensor spoof detection performance from 67.60\% to 80.63\% when tested on the LivDet 2017 dataset. Training the UMG wrapper requires only 100 live fingerprint images from the target sensor, alleviating the time and resources required to generate large-scale live and spoof datasets for a new sensor. We also fabricate physical spoof artifacts using a mixture of known spoof materials to explore the role of cross-material style transfer in improving generalization performance.
\end{abstract}

\begin{IEEEkeywords}
Fingerprint spoof detection, presentation attack detection, liveness detection, generalization, style transfer, fingerprint spoof buster
\end{IEEEkeywords}

\IEEEpeerreviewmaketitle

\section{Introduction}
\label{sec:introduction}

\begin{figure}[t]
\centering
\includegraphics[trim=1cm 0.2cm 1.1cm 0cm, width=0.82\linewidth]{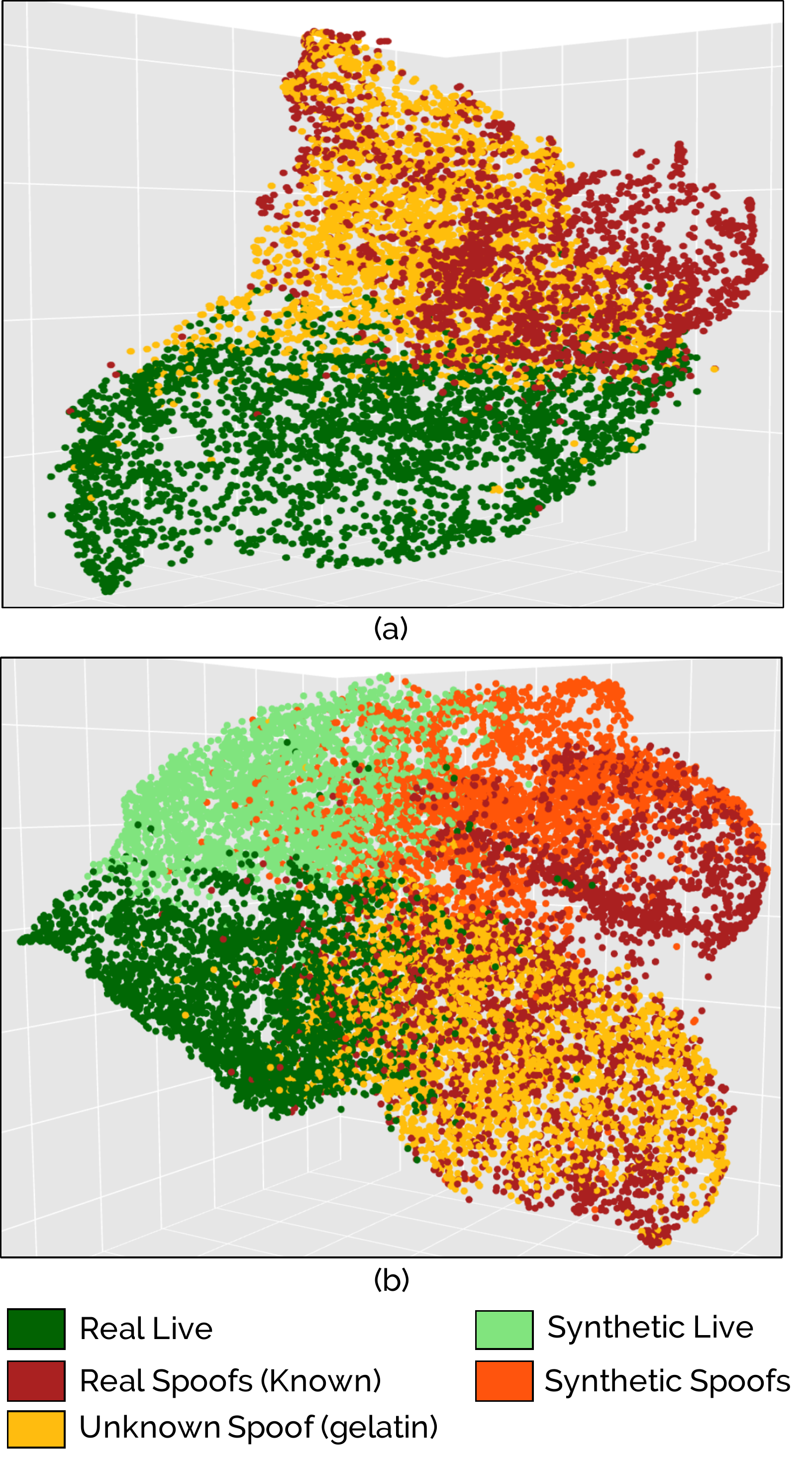}
\caption{3D t-SNE visualization of feature embeddings learned by Fingerprint Spoof Buster~\cite{chugh2018fingerprint} of (a) live (dark green) and eleven known spoof materials (red) {\scriptsize\textit{(2D printed paper, 3D universal targets, conductive ink on paper, dragon skin, gold fingers, latex body paint, monster liquid latex, play doh, silicone, transparency, and wood glue)}} used in training, and unknown spoof, gelatin (yellow). A large overlap between unknown spoof (gelatin) and live feature embeddings indicate poor generalization performance of state of the art spoof detector. (b) Synthetic live (bright green) and synthetic spoof (orange) images generated by the proposed Universal Material Generator (UMG) wrapper improve the separation between real live and real spoof. 3D t-SNE visualizations are available at http://tarangchugh.me/posts/umg/index.html}
\label{fig:fig1}
\vspace{-2mm}
\end{figure}


\IEEEPARstart{W}{ITH} the proliferation of automated fingerprint recognition systems in many applications, including mobile payments, international border security, and national ID, fingerprint spoof attacks are of increasing concern~\cite{marcel2019handbook, IARPAProject}. Fingerprint spoof attacks, one of the most common forms of presentation attacks\footnote{The ISO standard \textit{IEC 30107-1:2016(E)}~\cite{isopad} defines presentation attacks as the \textit{``presentation to the biometric data capture subsystem with the goal of interfering with the operation of the biometric system".}}, include the use of~\textit{gummy fingers}~\cite{matsumoto2002impact} and~\textit{2D or 3D printed fingerprint targets}~\cite{cao2016hacking, arora2016design, arora2017goldfingers, engelsma2018universal}, \textit{i.e.} fabricated finger-like objects with an accurate imitation of one's fingerprint to steal their identity. Other forms of presentation attacks include use of~\textit{altered fingerprints}~\cite{yoon2012altered, tabassi2018altered}, \textit{i.e.} intentionally tampered or damaged real fingerprint patterns to avoid identification, and~\textit{cadaver fingers}~\cite{marasco2015survey}.



\begin{figure*}[htbp!]
\centering
\includegraphics[trim=0cm 0.8cm 0cm 0cm, width=\linewidth]{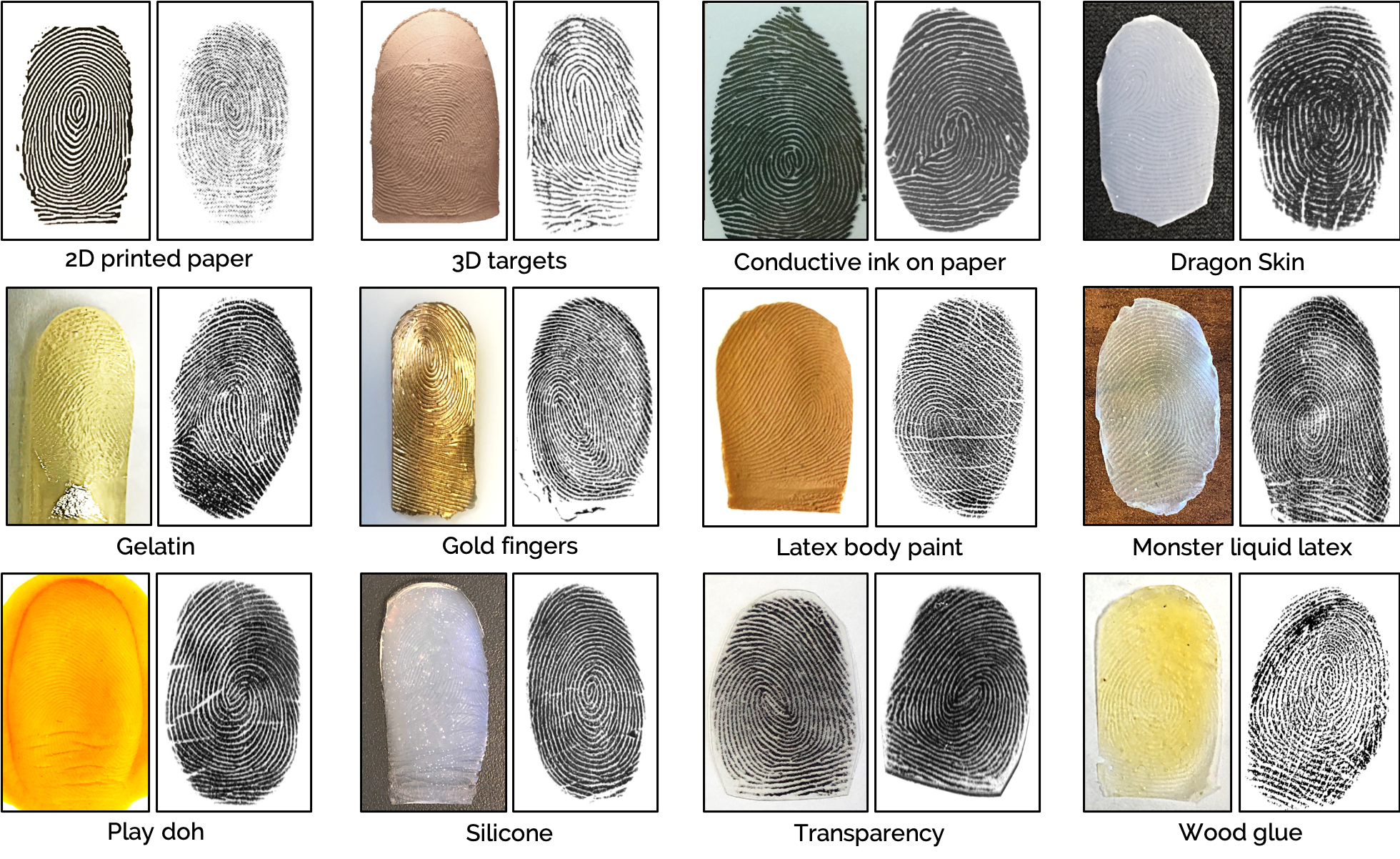}
\caption{Illustration of physical spoof artifacts and the corresponding images. Spoofs are fabricated using twelve different readily available and inexpensive spoof materials. The physical artifacts and their fingerprint images do not necessarily correspond to the same finger.}
\label{fig:spoofs}
\end{figure*}

Fingerprint spoof attacks can be realized using a multitude of fabrication processes ranging from basic \textit{molding and casting} to utilizing sophisticated 2D and 3D printing techniques~\cite{matsumoto2002impact, cao2016hacking, arora2017goldfingers}. Readily available and inexpensive materials such as gelatin, play doh, and wood glue, have been utilized to fabricate high fidelity fingerprint spoofs which are capable of bypassing a fingerprint recognition system. For example, in March 2013, a Brazilian doctor was arrested for using spoof fingers made of silicone to fool the biometric attendance system at a hospital in Sao Paulo\footnote{https://www.bbc.com/news/world-latin-america-21756709}. In July 2016, researchers at Michigan State University unlocked a fingerprint secured-smartphone using a 2D printed fingerprint spoof to help police with a homicide case\footnote{http://statenews.com/article/2016/08/how-msu-researchers-unlocked-a-fingerprint-secure-smartphone-to-help-police-with-homicide-case}, using the technique proposed in~\cite{cao2016hacking}. In March 2018, a gang in Rajasthan, India, was arrested for spoofing the biometric attendance system, using glue casted in wax molds, to provide proxies for a police entrance exam\footnote{https://www.medianama.com/2018/03/223-cloned-thumb-prints-used-to-spoof-biometrics-and-allow-proxies-to-answer-online-rajasthan-police-exam/}. As recent as April 2019, a Galaxy S10 owner with a 3D printer and a photo of his own fingerprint was able to spoof the ultrasonic in-display fingerprint sensor on his smartphone\footnote{https://imgur.com/gallery/8aGqsSu}. Other similar successful spoof attacks have been reported showing the vulnerabilities of fingerprint biometric systems\footnote{http://fortune.com/2016/04/07/guy-unlocked-iphone-play-doh/}$^,$\footnote{https://srlabs.de/bites/spoofing-fingerprints/}. It is likely that a large number of these attacks are never detected and hence not reported.
In response to this growing threat, a series of fingerprint Liveness Detection (LivDet) competitions~\cite{yambay2019review} have been held since 2009 to benchmark various spoof detection solutions. See~\cite{orru2019livdet} for results of the LivDet 2019. Another initiative is the IARPA ODIN Program~\cite{IARPAProject} with the goal of developing robust spoof detection systems for fingerprints, face, and iris biometric modalities.

Generally, fingerprint spoofs can be detected by either (i) hardware-based, or (ii) software-based approaches~\cite{marcel2019handbook, marasco2015survey}. In the case of hardware-based approaches, the fingerprint readers are augmented with sensor(s) which detect characteristics of vitality, such as blood flow, thermal output, heartbeat, skin distortion, and odor~\cite{antonelli2006fake, baldisserra2006fake}. Additionally, special types of fingerprint sensing technologies have been developed for imaging the sub-dermal friction ridge surface based on multi-spectral~\cite{robison2019system}, short-wave infrared~\cite{tolosana2018towards} and optical coherent tomography (OCT)~\cite{moolla2019optical, chugh2019oct}. An open-source fingerprint reader, called RaspiReader, uses two cameras to provide complementary streams (direct-view and FTIR) of images for spoof detection~\cite{engelsma2017raspireader}. Ultrasound-based in-display fingerprint readers developed for smartphones by Qualcomm Inc.~\cite{agassy2019liveness} utilize acoustic response characteristics for spoof detection.

\begin{table*}[!htbp]
\caption{Summary of the studies primarily focused on fingerprint spoof generalization.}
\label{tab:litrev}
\centering
\resizebox{\linewidth}{!}{
\begin{tabular}{  p{2.8 cm}  >{\centering\arraybackslash}p{6 cm}  >{\centering\arraybackslash}p{3.3cm}  >{\centering\arraybackslash}p{4.4cm}  }
\toprule
\textbf{Study} & \textbf{Approach} & \textbf{Database} & \textbf{Performance} \\ \bottomrule

Rattani et al.~\cite{rattani2015open} &
Weibull-calibrated SVM &
LivDet 2011 &
EER = 19.70\%\\ \midrule

Ding \& Ross~\cite{ding2016ensemble} &
Ensemble of multiple one-class SVMs &
LivDet 2011&
EER = 17.60\%\\ \midrule

Chugh \& Jain~\cite{chugh2018fingerprint} &
MobileNet trained on minutiae-centered local patches &
LivDet 2011-2015 &
ACE = 1.48\% (LivDet 2015), 2.93\% (LivDet 2011, 2013)\\ \midrule

Chugh \& Jain~\cite{chugh2019fingerprint} &
Identify a representative set of spoof materials to cover the deep feature space &
MSU-FPAD v2.0, 12 spoof materials &
TDR = 75.24\% @ FDR = 0.2\% \\ \midrule

Engelsma \& Jain~\cite{engelsma2019generalizing} & 
Ensemble of generative adversarial networks (GANs) &
Custom database with live and 12 spoof materials &
TDR = 49.80\% @ FDR = 0.2\% \\ \midrule

González-Soler et al.~\cite{gonzalez2019fingerprint} &
Feature encoding of dense-SIFT features &
LivDet 2011-2015 &
TDR =  7.03\% @ FDR = 1\% (LivDet 2015), ACE = 1.01\% (LivDet 2011, 2013)\\ \midrule

Tolosana et al.~\cite{tolosana2019biometric} &
Fusion of two CNN architectures trained on SWIR images &
Custom database with live and 8 spoof materials &
EER = 1.35\%\\ \midrule

Gajawada et al.~\cite{gajawada2019universal} \hspace{2mm} (preliminary work) &
Style transfer from spoof to live images to improve generalization; requires few samples of target material &
LivDet 2015, CrossMatch sensor &
TDR = 78.04\% @ FDR = 0.1\% \\ \toprule

\textbf{Proposed Approach} &
Style transfer between known spoof materials to improve generalizability against completely unknown materials &
MSU-FPAD v2.0, 12 spoof materials \& LivDet 2017 & 
TDR = \textbf{91.78\%} @ FDR = 0.2\% (MSU-FPAD v2.0); Avg. Accuracy = \textbf{95.88\%} (LivDet 2017) \\ \bottomrule

\end{tabular}
}
\flushleft ACE = Average Classification Error; EER = Equal Error Rate; TDR = True Detection Rate (spoofs); FDR = False Detection Rate (spoofs)
\end{table*}


In contrast, software-based solutions extract salient features from the captured fingerprint image (or a sequence of frames) for separating live and spoof images. The software-based approaches in the literature are typically based on (i) anatomical features (e.g. pore locations and their distribution~\cite{schuckers2017fingerprint}), (ii) physiological features (e.g. perspiration~\cite{marasco2012combining}), and (iii)~texture-based features (e.g. Weber Local Binary Descriptor (WLBD)~\cite{xia2018novel}, SIFT~\cite{gonzalez2019fingerprint}. Most state-of-the-art approaches are learning-based, where the features are learned by training convolutional neural networks (CNN)~\cite{nogueira2016fingerprint, jang2017fingerprint, chugh2017fingerprint, pala2017deep, tolosana2018towards, chugh2018fingerprint}.


\begin{figure*}[htbp!]
\centering
\includegraphics[trim=0cm 1cm 0cm 0cm, width=\linewidth]{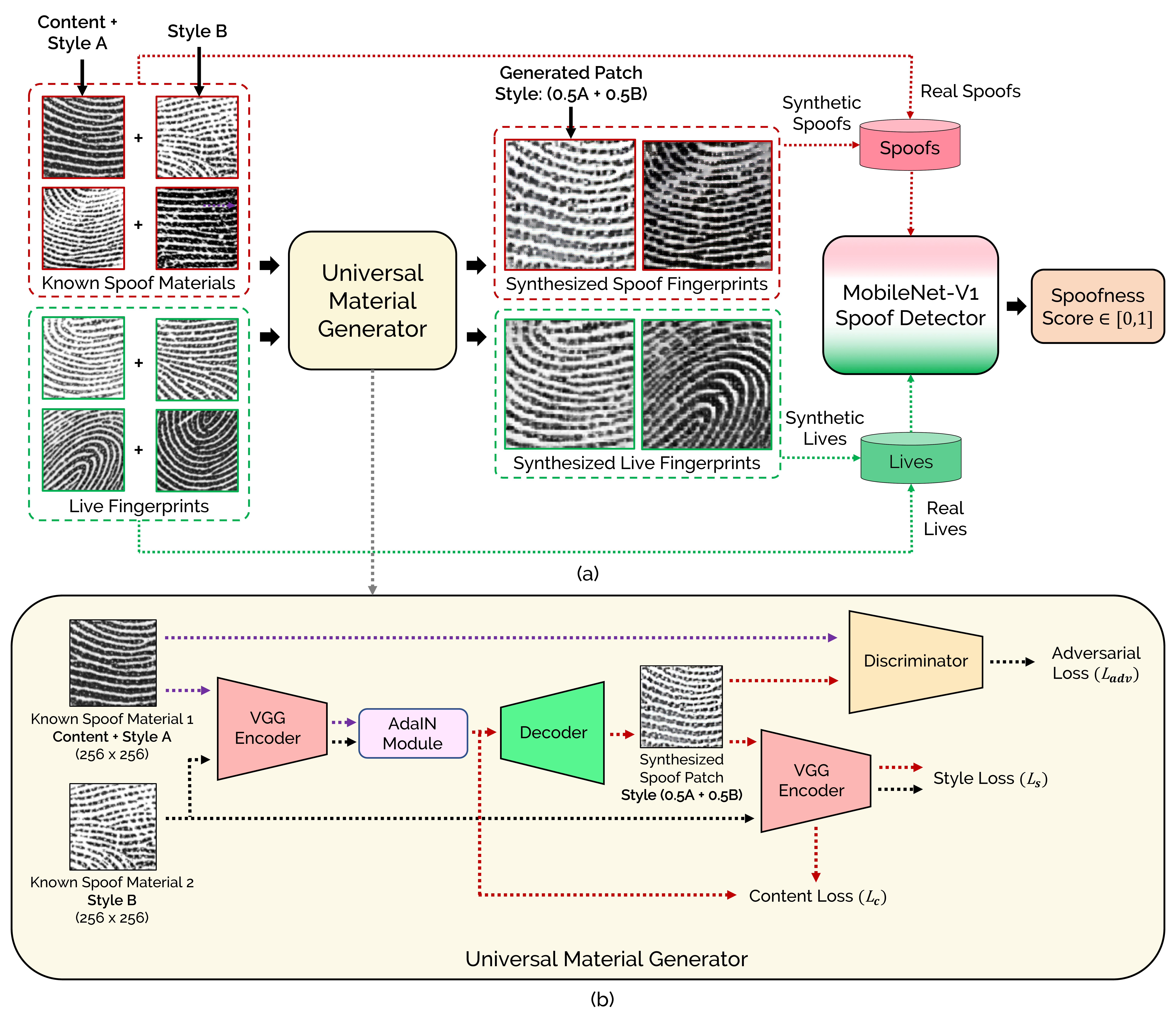}
\caption{Proposed approach for (a) synthesizing spoof and live fingerprint patches, and (b) design of the proposed Universal Material Generator (UMG) wrapper. An AdaIN module is used for performing the style transfer in the encoded feature space. The same VGG-19~\cite{simonyan2014very} encoder is used for computing content loss and style loss. A discriminator similar to the one used in DC-GAN~\cite{radford2015unsupervised} is used for computing the adversarial loss. The synthesized patches can be used to train any fingerprint spoof detector. Hence, our approach is referred to as a wrapper which can be used in conjunction with any spoof detector.}
\label{fig:overview}
\end{figure*}


One of the major limitations of current spoof detection methods is their poor generalization performance across ``unknown" spoof materials, that were not used during training of the spoof detector. To generalize an algorithm's effectiveness across spoof fabrication materials, called \textit{cross-material} performance, spoof detection has been referred to as an \textit{open-set problem}\footnote{Open-set problems address the possibility of new classes during testing, that were not seen during training. Closed-set problems, on the other hand, evaluate only those classes that the system was trained on.}~\cite{rattani2015open}. Table~\ref{tab:litrev} presents a summary of the studies primarily focused on generalization. Engelsma and Jain~\cite{engelsma2018raspireader, engelsma2019generalizing} proposed using an ensemble of generative adversarial networks (GANs) on live fingerprint images with the hypotheses that features learned by a discriminator to distinguish between real live and synthesized live fingerprints can be used to separate live fingerprints from spoof fingerprints as well. One limitation of this approach is that the discriminator in the GAN architecture may learn many features related to structural noise added by the generative process. Such features are likely not present in the spoofs fabricated with unknown materials. 



It has been shown that the selection of spoof materials used in training (known spoofs) directly impacts the performance against unknown spoofs~\cite{rattani2015open, chugh2019fingerprint}. In particular, Chugh and Jain~\cite{chugh2019fingerprint} analyzed the material characteristics (two optical and two physical) of 12 different spoof materials to identify a representative set of six materials that cover most of the spoof feature space. Although, this approach can be used to identify if including a new spoof material in training dataset would be beneficial, it does not improve the generalization performance against materials that are unknown during training. With the increasing popularity of fingerprint authentication systems, hackers are constantly devising new fabrication techniques and novel materials to attack them. It is prohibitively expensive to include all spoof fabrication materials in training a spoof detector.

Additionally, fingerprint images captured using different fingerprint sensors, typically, have unique characteristics due to different sensing technologies, sensor noise, and varying resolution. As a result, fingerprint spoof detectors, especially CNN-based, are known to suffer from poor generalization performance in the cross-sensor scenario, where the spoof detector is trained on images captured using one sensor and tested on images from another. Improving cross-sensor spoof detection performance is important in order to alleviate the time and resources involved in collecting large-scale datasets with the introduction of new sensors.

In this paper, we propose a style-transfer based method to improve the cross-material and cross-sensor generalization performance of fingerprint spoof detectors. In particular, for the cross-material scenario, we hypothesize that the texture (style) information from the known spoof fingerprint images can be transferred from one spoof type to another type to synthesize spoof images potentially similar to spoofs fabricated from materials not seen in the training set. In the cross-sensor scenario, we utilize a small set of live fingerprint images $(\sim100)$ from the target sensor, say Green Bit, to transfer its sensor-specific style characteristics to large-scale live and spoof datasets available from a source sensor, say Digital Persona. Our framework, called \textit{Universal Material Generator} (UMG), is used to augment CNN-based spoof detectors, significantly improving their performance against novel materials, while retaining their performance on known materials. See Figure~\ref{fig:umg_output_spoofs} for examples of some of the style transferred images.


Realistic image synthesis is a challenging problem. Early non-parametric methods faced difficulty in generating images with textures that are not known during training~\cite{chen2009sketch2photo}. Machine learning has been very effective in this regard, both in terms of realism and generality. Gatys et al.~\cite{gatys2015neural} perform artistic style transfer, combining the content of an image with the style of any other by minimizing the feature reconstruction loss and a style reconstruction loss which are based on features extracted from a pre-trained CNN at the same time. While this approach generates realistic looking images, it is computationally expensive since each step of the optimization requires a forward and backward pass through the pre-trained network. Other studies~\cite{johnson2016perceptual, wang2017multimodal, li2016precomputed} have explored training a feed-forward network to approximate solutions to this optimization problem. There are other methods based on feature statistics to perform style transfer~\cite{ulyanov2017improved, huang2017arbitrary}. Elgammal et al.~\cite{elgammal2017can} applied GANs to generate artistic images. Isola et al.~\cite{isola2017image} used conditional adversarial networks to learn the loss for image-to-image translation. Xian et al.~\cite{xian2018texturegan} learnt to synthesize objects consistent with texture suggestions. The proposed Universal Material Generator builds on~\cite{huang2017arbitrary} and is capable of producing realistic fingerprint images containing style (texture) information from images of two different spoof materials. Existing style transfer methods condition the source image with target material style. However, in the context of fingerprint synthesis, this results in a loss in fingerprint ridge-valley information (\textit{i.e.} content). In order to preserve both style and content, we use adversarial supervision to ensure that the synthesized images appear similar to the real fingerprint images.

The main contributions of this study are enumerated below.

\begin{itemize}


\item A style-transfer based wrapper, called Universal Material Generator (UMG), to improve the generalization performance of any fingerprint spoof detector against spoofs made from materials not seen during training. It attempts to synthesize impressions with style (texture) characteristics potentially similar to unknown spoof materials by interpolating the styles from known spoof materials.



\item Experiments on a database of $5,743$ live and $4,912$ spoof images of $12$ different materials to demonstrate that the proposed approach improves the cross-material generalization performance of a state-of-the-art spoof detector from TDR of 75.4\% to TDR of 91.78\% @ FDR = 0.2\%. Additionally, experimental results on LivDet 2017 datasets show that the proposed approach achieves state-of-the-art performance.

\item Improved the cross-sensor spoof detection performance by synthesizing large-scale live and spoof datasets using only 100 live images from a new target sensor. Our approach is shown to improve the average cross-sensor spoof detection performance from 67.60\% to 80.63\% on LivDet 2017 dataset. 


\item Used 3D t-SNE visualization to interpret the performance improvement against unknown spoof materials. 

\item Fabricated physical spoof artifacts using a mixture of known spoof materials to show that the synthetically generated images using fingerprint images of the same set of spoof materials correspond to an unknown material with similar style (texture) characteristics.

\end{itemize}

Our preliminary work~\cite{gajawada2019universal} utilized a few impressions of a known spoof material to generate more impressions of that material. It improved the spoof detection performance against ``known" spoof materials for which only limited training data is available. In comparison, the proposed approach interpolates the style characteristics of known spoof materials to improve the spoof detection performance against ``unknown" spoof materials. The proposed approach is also shown to improve the cross-sensor generalization performance.





\section{Proposed Approach}
The proposed approach includes three stages: (i)~training the Universal Material Generator (UMG) wrapper using the spoof images of known materials (with one material left-out from training), (ii)~generating synthetic spoof images using randomly selected image pairs of different but known materials, and (iii)~training a spoof detector on the augmented dataset to evaluate its performance on the ``unknown" material left out from training. In all our experiments, we utilize local image patches ($96 \times 96$) centered and aligned using minutiae location and orientation, respectively~\cite{chugh2018fingerprint}. During the evaluation stage, the spoof detection decision is made based on the average of spoofness scores for individual patches output from the CNN model.  An overview of the proposed approach is presented in Fig.~\ref{fig:overview}.

\begin{figure*}[htbp!]
\centering
\includegraphics[trim=0cm 0cm 0cm 0cm, width=\linewidth]{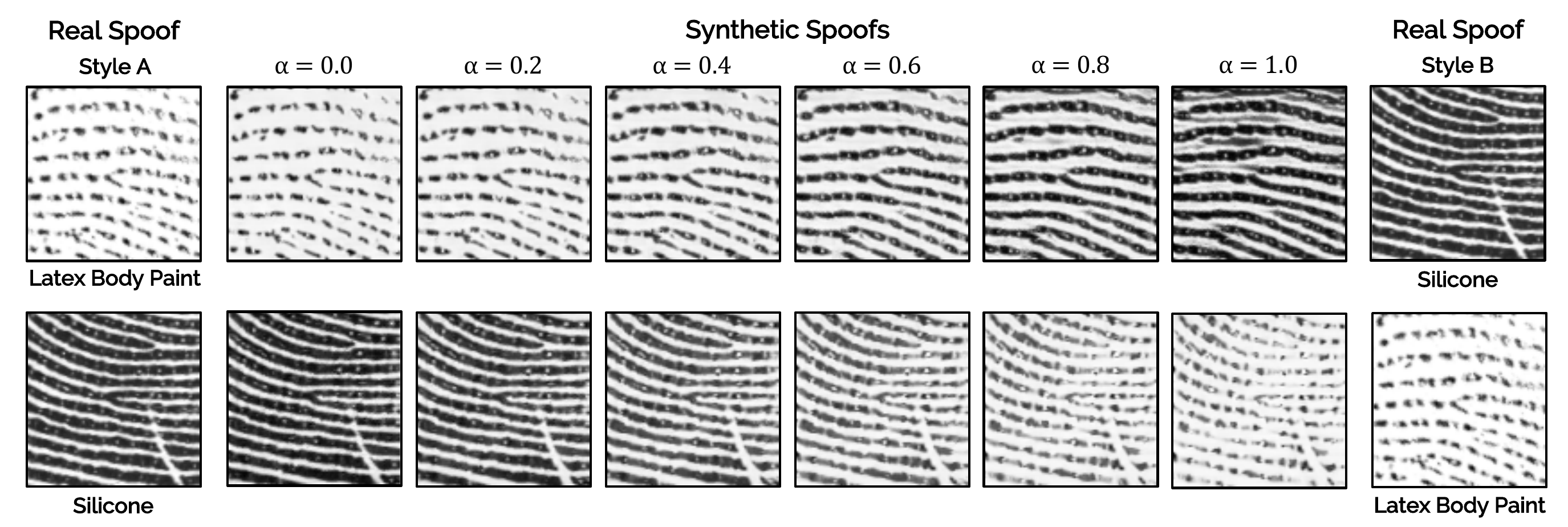}
\caption{Style transfer between real spoof patches fabricated with latex body paint and silicone to generate synthetic spoof patches using the proposed Universal Material Generator (UMG) wrapper. The extent of style transfer can be controlled by the parameter $\alpha \in [0,1]$.}
\label{fig:style_interpolation}
\end{figure*}

\subsection{Universal Material Generator (UMG) Wrapper}
\label{sec:umg}
The primary goal of the UMG wrapper is to generate synthetic spoof images corresponding to unknown spoof materials, by transferring the style (texture) characteristics between fingerprint images of known spoof materials. Gatys et al.~\cite{gatys2016image} were the first to show that deep neural networks (DNNs) could encode not only content but also the style information. They proposed an optimization-based style-transfer approach, although prohibitively slow, for arbitrary images. 
In~\cite{ulyanov2017improved}, Ulyanov et al. proposed use of an InstanceNorm layer to normalize feature statistics across spatial dimensions. An InstanceNorm layer is designed to perform the following operation:

\begin{equation} \label{eqn:basic} IN(x) = \gamma\Big(\dfrac{x - \mu(x)}{\sigma(x)}\Big) + \beta \end{equation} 

where, $x$ is the input feature space, $\mu(x)$ and $\sigma(x)$ are the mean and standard deviation parameters, respectively, computed across spatial dimensions independently for each channel and each sample. It was observed that changing the affine parameters $\gamma$ and $\beta$ (while keeping convolutional parameters fixed) leads to variations in the style of the image, and the affine parameters could be learned for each particular style. This motivated an approach for artistic style transfer~\cite{dumoulin2016learned}, which learns $\gamma$ and $\beta$ values for each feature space and style pair. However, this required retraining of the network for each new style. 

Huang and Belongie~\cite{huang2017arbitrary} replaced the InstanceNorm layer with an Adaptive Instance Norm (AdaIN) layer, which can directly compute affine parameters from the style image, instead of learning them -- effectively transferring style by imparting second-order statistics from the target style image to the source content image, through the affine parameters. We follow the same approach as described in \cite{huang2017arbitrary} in UMG wrapper for fusing feature statistics of one known (source) spoof material image ($c$) providing friction ridge (content) information and source style, with another known, but different (target style) spoof material ($s$) in the feature space. As described in AdaIN, we apply instance normalization on the input source image feature space however not with learnable affine parameters. The channel-wise mean and variance of the source image's feature space is aligned to match those of the target image's feature space. This is done by computing the affine parameters from the target material spoof feature space in the following manner:
\begin{equation} 
AdaIN(x, y) = \sigma(y)\Big(\dfrac{x - \mu(x)}{\sigma(x)}\Big) + \mu(y)
\label{eq:adain}
\end{equation}

where the source ($c$) feature space is $x$ and the target ($s$) feature space is $y$. In this manner, $x$ is normalized with $\sigma(y)$ and shifted by $\mu(y)$. Our synthetic spoof generator $G$ is composed of an encoder $f(\cdot)$ and a decoder $g(\cdot)$. For the encoder, $f(\cdot)$, we use the first few layers of a pre-trained VGG-19 network similar to~\cite{johnson2016perceptual}. The weights of this network are frozen throughout all stages of the setup. For source image $(c)$ and the target image $(s)$, $x$ is $f(c)$ and $y$ is $f(s)$. The desired feature space is obtained as: 
\begin{equation}
t = AdaIN(f(c), f(s))
\end{equation}

We use the decoder, $g(\cdot)$, to take $t$ as input to produce $T(c,s) = g(t)$ which is the final synthesized image conditioned on the style from the target image. In order to ensure that our synthesized spoof patches i.e $g(t)$ do match the style statistics of the target material spoof, we apply a style loss $L_s$ similar to~\cite{johnson2016perceptual, li2017demystifying} given as:
\begin{equation} 
\begin{array} { r } { \mathcal { L } _ { s } = \sum _ { i = 1 } ^ { L } \left\| \mu \left( \phi _ { i } ( g ( t ) ) \right) - \mu \left( \phi _ { i } ( s ) \right) \right\| _ { 2 } + }\\ { \sum _ { i = 1 } ^ { L } \left\| \sigma \left( \phi _ { i } ( g ( t ) ) \right) - \sigma \left( \phi _ { i } ( s ) \right) \right\| _ { 2 } } \end{array} 
\label{eq:styleloss}
\end{equation}




where each $\phi_i$ denotes a layer in the VGG-19 network we use as encoder. We pass $g(t)$ and $s$ through $f(\cdot)$ and extract the outputs of $relu1\_1$, $relu2\_1$, $relu3\_1$ and $relu4\_1$ layers for computing $\mathcal{L}_s$.

The extent of style transfer can be controlled by interpolating between feature maps that are:
\begin{equation}
T(c,s,\alpha) = g((1-\alpha) \cdot f(c) + \alpha \cdot t)
\end{equation}

where setting $\alpha=0$ will reconstruct the original content image and $\alpha=1$ will construct the most stylized image. To combine the two known styles, we preserve the style of source spoof material while conditioning it with target spoof material by setting the value of $\alpha$ to $0.5$.

\begin{algorithm}[t]
\caption{Training UMG wrapper}
\begin{algorithmic}[1]
\Procedure{}{}
\BState \emph{input}
\State $x$: source image providing friction ridge content and known style A
\State $y$: target image providing known style B
\State $f(\cdot)$: encoder network; first 4 layers of VGG-19 network pre-trained on ImageNet with weights frozen during training
\State $g(\cdot)$: decoder network; mirrors $f(\cdot)$ with pooling layers replaced with nearest up-sampling layers
\State $D(\cdot)$: discriminator function similar to~\cite{radford2015unsupervised} 
\State $A(x, y)$: AdaIN operation; transfer style from $x$ to $y$ (using Eq. \ref{eq:adain})
\State $\alpha = 0.5$
\State $\lambda_c = 0.001$, $\lambda_s = 0.002$
\BState \emph{output}
\State $UMG(\cdot)$: UMG wrapper trained on known materials

\BState \emph{begin}:
\State \emph{Encoding}: $f_x = f(x)$ and $f_y = f(y)$
\State \emph{Style transfer}:  $t = A(f_x, f_y)$
\State \emph{Stylized image}: $T(c,s,\alpha) = g((1-\alpha) \cdot f_c + \alpha \cdot t)$
\State \emph{Style Loss}: $\mathcal{L}_s$ using Eq.~\ref{eq:styleloss}
\State \emph{Content Loss}: $\mathcal{L}_c$ using Eq.~\ref{eq:contentloss}
\State \emph{Adversarial Loss (generator)}: $\mathcal{L}^G_{adv}$ using Eq.~\ref{eq:adv_gen}
\State \emph{Adversarial Loss (discriminator)}: $\mathcal{L}^D_{adv}$ using Eq.~\ref{eq:adv_dis}
\State \emph{Objective functions for training UMG wrapper}
\State  $\min_G \mathcal{L}_G = \lambda_c \cdot \mathcal{L}_c + \lambda_s \cdot \mathcal{L}_s + \mathcal{L}^G_{adv}$
\State $ \max_D \mathcal{L}_D = \mathcal{L}^D_{adv}$
\BState \emph{end}
\EndProcedure
\end{algorithmic}
\label{algo:umg}
\end{algorithm}

\begin{figure*}[htbp!]
\centering
\includegraphics[trim=0.7cm 0.4cm 0.5cm 0.5cm, width=\linewidth]{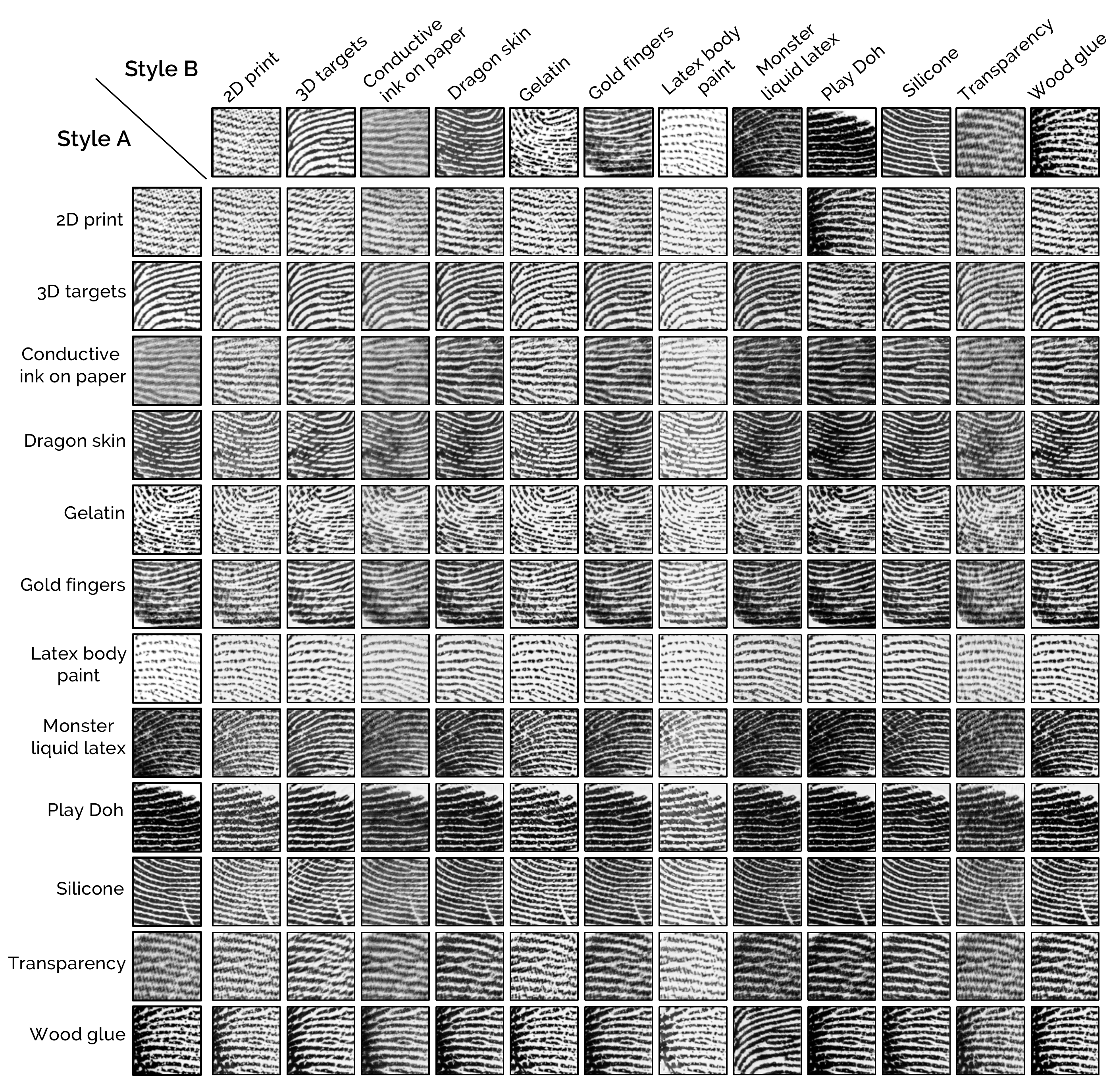}
\caption{Synthesized spoof patches (96 x 96) by the proposed Universal Material Generator using patches of a known (source) material (first column) conditioned on style ($\alpha=0.5$) of another (target) known material (first row).}
\label{fig:umg_output_spoofs}
\end{figure*}

To ensure that the synthesized images retain friction ridge (fingerprint) content from the real image, we use a content loss, $\mathcal{L}_c$, which is computed as the euclidean distance between the features of the synthesized image \textit{i.e.} $f(g(t))$ and the target features ($t$) from the real image.
\begin{equation}
\mathcal{L}_c  = \left\| f(g(t)) - t \right\| _ { 2 }
\label{eq:contentloss}
\end{equation}

Doing the style transfer, simply using a content loss $(\mathcal{L}_c)$ to ensure that content is retained is not enough to ensure that the synthesized images look like real images. Fingerprints have many details in terms of structure due to the presence of certain landmarks e.g. minutiae, ridges, and pores. With the aim of synthesizing fingerprints that look indistinguishable from the real fingerprints, we use adversarial supervision. A typical generative adversarial network (GAN) setup consists of a generator $G$ and a discriminator $D$ playing a \textit{minimax game}, where $D$ tries to distinguish between synthesized and real images, and $G$ tries to fool $D$ by generating realistic looking images. The adversarial objective functions for the generator ($\mathcal{L}_{adv}^G$) and discriminator ($\mathcal{L}_{adv}^D$) are given as\footnote{Here $x$ is an image sampled from the distribution of real fingerprints, and $t$ is the feature output by the AdaIN module.}:
\begin{equation}
\mathcal{L}_{adv}^G = \mathbb{E}_{t} [\log (1 - D(G(t)))]
\label{eq:adv_gen}
\end{equation}

\begin{equation}
\mathcal{L}_{adv}^D = \mathbb{E}_{x} [\log D(x)] + \mathbb{E}_{t} [\log (1-D(G(t)))]
\label{eq:adv_dis}
\end{equation}

In our approach, we use a discriminator as used in~\cite{radford2015unsupervised} and the generator is the decoder function $g(\cdot)$. We optimize the UMG wrapper in an end-to-end manner with the following objective functions:
\begin{equation}
 \min_G \mathcal{L}_G = \lambda_c \cdot \mathcal{L}_c + \lambda_s \cdot \mathcal{L}_s + \mathcal{L}_{adv}^G
\end{equation}
\begin{equation}
 \max_D \mathcal{L}_D = \mathcal{L}_{adv}^D
\end{equation}

where $\lambda_c$ and $\lambda_s$ are the weight parameters for content loss $(\mathcal{L}_c)$ and style loss $(\mathcal{L}_s)$, respectively. Algorithm~\ref{algo:umg} summarizes the steps involved in training a UMG wrapper.

\begin{figure}[t]
\centering
\includegraphics[trim=0.2cm 0.2cm 0.2cm 0.2cm, width=\linewidth]{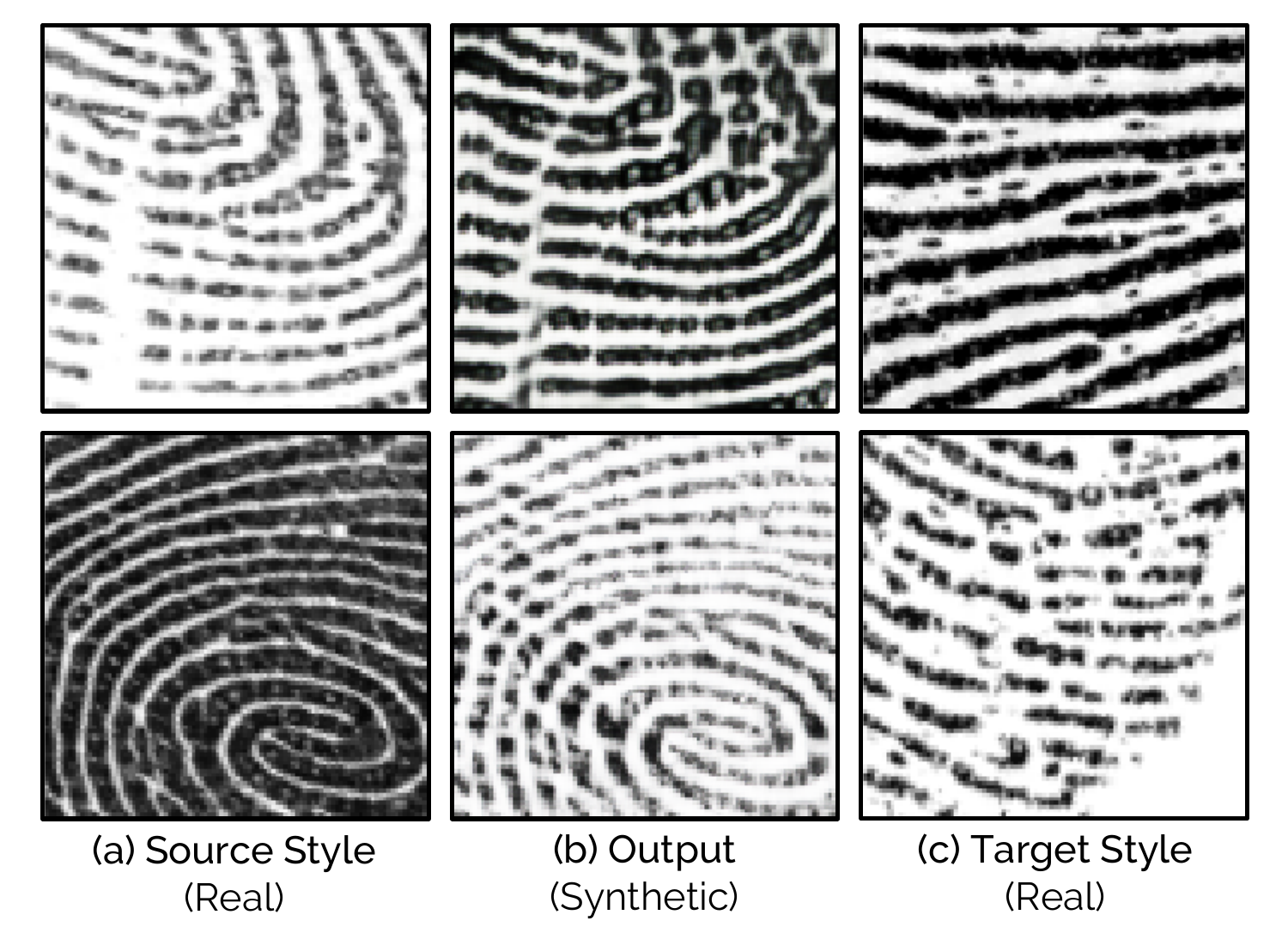}
\caption{Synthetic live images generated by the proposed Universal Material Generator. (a) Source style images, (c) target style images, and (b) synthesized live images.}
\label{fig:lives_synth}
\end{figure}

\begin{table*}[t]
\centering
\caption{Summary of the MSU-FPAD v2 and LivDet 2017 datasets.}
\label{tab:database}
\resizebox{\linewidth}{!}{
\begin{tabular}{ p{3.5 cm} >{\centering\arraybackslash}p{6.9 cm} >{\centering\arraybackslash}p{2.1 cm} >{\centering\arraybackslash}p{2.1 cm} >{\centering\arraybackslash} p{2.1 cm} }
\toprule
\textbf{Dataset} & \textbf{MSU-FPAD v2~\cite{chugh2019fingerprint}} & \multicolumn{3}{c}{\textbf{LivDet 2017~\cite{mura2018livdet}}} \\ \toprule

\multirow{2}{*}{\textbf{Fingerprint Reader}} & \multirow{2}{*}{CrossMatch Guardian 200} & GreenBit & Orcanthus  & Digital Persona \\ 

 &  & Dacty Scan 84C & Certis2 Image & U.are.U 5160 \\ \midrule

\textbf{Image Size ($px.$) ($w \times h$)} & $ 800 \times 750 $ & $ 500 \times 500 $ & $300 \times n^{\dag}$ & $252 \times 324$ \\ \midrule

\textbf{Resolution ($dpi$)} & $500$ & $569$ & $500$ & $500$ \\ \midrule

\textbf{\#Live Images (Train / Test)} & $4,743$ / $1,000$ & $1,000$ / $1,700$ & $1,000$ / $1,700$  & $999$ / $1,692$ \\ \midrule

\textbf{\#Spoof Images (Train / Test)} & $4,912$ (leave-one-out) & $1,200$ / $2,040$ & $1,180^*$ / $2,018$ & $1,199$ / $2,028$ \\ \midrule

\textbf{Known Spoof Materials (Training)} & \multirow{2}{*}{\parbox{6.5cm}{\vspace{1mm}\textbf{Leave-one-out}: 2D Printed Paper, 3D Universal Targets, Conductive Ink on Paper, Dragon Skin, Gelatin, \\Gold Fingers, Latex Body Paint, Monster Liquid Latex, Play Doh, Silicone, Transparency, Wood Glue}} & \multicolumn{3}{>{\centering\arraybackslash}p{7.5cm}}{Wood Glue, Ecoflex, Body Double} \\ 

\textbf{Unknown Spoof Materials (Testing)} & & \multicolumn{3}{>{\centering\arraybackslash}p{7.5cm}}{Gelatine, Latex, Liquid Ecoflex} \\ \bottomrule
\end{tabular}
}
\flushleft $\dag$ Fingerprint images captured using Orcanthus reader have a variable height ($350-450 px$) depending on the friction ridge content.\\
*A set of $20$ Latex spoof fingerprints found in the training set of Orcanthus fingerprint reader were excluded in our experiments. Only Wood Glue, Ecoflex, and Body Double are expected to be in the training dataset.
\end{table*}

\subsection{UMG-Wrapper for Spoof Generalization}

Given a spoof dataset of real images, $S^m_{real}$, fabricated using a set of $m$ spoof materials, we adopt a leave-one-out protocol to split the dataset such that spoof images fabricated using $m-1$ materials are considered as ``known" and used for training. And the images fabricated using the left-out $m^{th}$ material are considered as ``unknown" and used for computing the generalization performance. The fingerprint images of known materials ($k = m-1$) are used to train the UMG wrapper (UMG$_{spoof}$) described in section~\ref{sec:umg}.

After we train the UMG$_{spoof}$, we utilize a total of $N_{synth}$ randomly selected pairs of images $\{I^i_{m_a}$, $I^i_{m_b}\}$ s.t. $i \in \{1, ..., N_{synth}\}$ from known but different materials $m_a, m_b \in \{m_1, ... m_k\}$, $a \ne b$,  to generate a dataset of synthesized spoof images $S^k_{synth}$. For each synthesized image, the friction ridge (content) information and the source material (style) characteristics are provided by the first image, $I_{m_a}$, and the target material (style) characteristics are provided by the second image, $I_{m_b}$. See Figures~\ref{fig:style_interpolation} and~\ref{fig:umg_output_spoofs}. The real spoof dataset is augmented with the synthesized spoof data to create a dataset that is used for training the fingerprint spoof detector. Additionally, we also augment the real live dataset with a total of $N_{synth}$ synthesized live images using another UMG wrapper (UMG$_{live}$) trained on only live images. Adding synthesized live data balances the data distribution and forces the spoof detector to learn generative-noise invariant features to distinguish between lives and spoofs. Figure~\ref{fig:lives_synth} presents examples of the synthesized live images.


\subsection{Fingerprint Spoof Detection}
The proposed Universal Material Generator approach acts like a wrapper on top of any existing spoof detector to make it more robust to spoofs not seen during training. In this study, we employ Fingerprint Spoof Buster~\cite{chugh2018fingerprint}, a state-of-the-art CNN-based approach, that utilizes local patches ($96 \times 96$) centered and aligned around fingerprint minutiae to train MobileNet-v1~\cite{howard2017mobilenets} architecture. It achieved state-of-the-art performance on publicly available LivDet databases~\cite{yambay2019review} and exceeded the IARPA Odin Project~\cite{IARPAProject} requirement of True Detection Rate (TDR) of 97.0\% @ False Detection Rate (FDR) = 0.2\%.

\begin{table*}[t]
\centering
\caption{Generalization performance (TDR (\%) @ FDR = $0.2\%$) with leave-one-out method on MSU-FPAD v2 dataset. A total of twelve models are trained where the material left-out from training is taken as the ``unknown" material for evaluating the model.}
\label{tab:msuresults}
\resizebox{\linewidth}{!}{
\begin{tabular}{ >{\centering\arraybackslash}p{3.5cm} | >{\raggedleft\arraybackslash}p{1cm} | >{\raggedleft\arraybackslash}p{2cm} |>{\centering\arraybackslash}p{3.7cm} | >{\centering\arraybackslash}p{3.8cm} } \toprule

\multirow{3}{*}{\textbf{Unknown Spoof Material}} & \multirow{3}{*}{\textbf{\# Images}} &  \multirow{3}{*}{\textbf{\# Local Patches}} & \multicolumn{2}{c}{\textbf{Generalization Performance (TDR (\%) @ FDR = 0.2\%)}} \\ \cline{4-5}
& & & \textbf{Fingerprint Spoof Buster~\cite{chugh2019fingerprint}} & \textbf{Fingerprint Spoof Buster + UMG wrapper}
 \\ \toprule
\textbf{Silicone} 			& $1,160$ & $38,145$ 	& $67.62$ & $\textbf{98.64}$ \\ \midrule
\textbf{Monster Liquid Latex} 	& $882$ 	& $27,458$ 	& $94.77$ & $\textbf{96.24}$ \\ \midrule
\textbf{Play Doh} 			& $715$ 	& $17,602$ 	& $58.42$ & $\textbf{72.36}$ \\ \midrule
\textbf{2D Printed Paper} 		& $481$ 	& $7,381$ 	& $55.44$ & $\textbf{80.22}$ \\ \midrule
\textbf{Wood Glue} 			& $397$ 	& $12,681$ 	& $86.38$ & $\textbf{98.97}$ \\ \midrule
\textbf{Gold Fingers} 		& $295$ 	& $9,402$ 	& $88.22$ & $\textbf{88.59}$ \\ \midrule
\textbf{Gelatin} 				& $294$ 	& $10,508$ 	& $54.95$ & $\textbf{97.96}$ \\ \midrule
\textbf{Dragon Skin} 			& $285$ 	& $7,700$ 	& $97.48$ & $\textbf{100.00}$ \\ \midrule
\textbf{Latex Body Paint} 		& $176$ 	& $6,366$ 	& $76.35$ & $\textbf{89.72}$ \\ \midrule
\textbf{Transparency} 		& $137$ 	& $3,846$ 	& $95.83$ & $\textbf{100.00}$ \\ \midrule
\textbf{Conductive Ink on Paper} & $50$ 	& $2,205$ 	& $90.00$ & $\textbf{100.00}$ \\ \midrule
\textbf{3D Universal Targets} 	& $40$ 	& $1,085$ 	& $95.00$ & $\textbf{100.00}$ \\ \toprule

\textbf{Total Spoofs} 			& $\textbf{4,912}$ & $\textbf{144,379}$ & \multicolumn{2}{c}{Weighted mean* ($\pm$ weighted s.d.)} \\ \midrule

\textbf{Total Lives} 		& $\textbf{5,743}$ & $\textbf{228,143}$ & $\textbf{75.24 $\pm$ 15.21}$ & $\textbf{91.78 $\pm$ 9.43}$ \\ \bottomrule
\end{tabular}
}
\flushleft *The generalization performance for each spoof material is weighted by the number of images to produce the weighted mean and standard deviation.
\end{table*}

\begin{table*}[t]
\centering
\caption{Performance comparison between the proposed approach and state-of-the-art results~\cite{mura2018livdet} reported on LivDet 2017 dataset for \textbf{cross-material experiments} in terms of Average Classification Error (ACE) and TDR @ FDR = 1.0\%.}
\label{tab:livdetresults}
\resizebox{\linewidth}{!}{

\begin{tabular}{  p{2 cm}  >{\centering\arraybackslash}p{3.1cm}  >{\centering\arraybackslash}p{2cm}  >{\centering\arraybackslash}p{3.3cm}  >{\centering\arraybackslash}p{2cm}  >{\centering\arraybackslash}p{3.3cm}  } \toprule

\textbf{LivDet 2017} & \textbf{LivDet 2017 Winner~\cite{mura2018livdet}} & \multicolumn{2}{ c }{\textbf{Fingerprint Spoof Buster~\cite{chugh2019fingerprint}}} & \multicolumn{2}{ c }{\textbf{Fingerprint Spoof Buster + UMG wrapper}} \\ \midrule

& \textbf{ACE (\%)} 	& \textbf{ACE (\%)} 	& \textbf{TDR @ FDR = 1.0\%} & \textbf{ACE (\%)} 	& \textbf{TDR @ FDR = 1.0\%} \\ 
\toprule

Green Bit 		& 96.44 		& 96.68 	& 91.07 	& \textbf{97.42}	& \textbf{92.29} \\ \midrule
Orcanthus 	& \textbf{95.59} & 94.51 	& 66.59 	& 95.01		& 74.45 \\ \midrule
Digital Persona & 93.71 		& 95.12 	& 62.29  	& \textbf{95.20} & \textbf{75.47} \\ \toprule

\textbf{Mean $\pm$ s.d.} & 95.25 $\pm$ 1.40  & 95.44 $\pm$ 1.12 & 73.32 $\pm$ 15.52 & \textbf{95.88} $\pm$ \textbf{1.34} & \textbf{80.74} $\pm$ \textbf{10.02}  \\
\bottomrule
\end{tabular}
}
\end{table*}

\section{Experiments and Results}

\begin{table*}[t]
\centering
\caption{Cross-sensor fingerprint spoof generalization performance on LivDet 2017 dataset in terms of Average Classification Error (ACE) and TDR @ FDR~=~1.0\%.}
\label{tab:crosssensor}
\resizebox{\linewidth}{!}{

\begin{tabular}{  p{2.4 cm}  >{\centering\arraybackslash}p{2.4cm}  >{\centering\arraybackslash}p{2cm}  >{\centering\arraybackslash}p{3.3cm}  >{\centering\arraybackslash}p{2cm}  >{\centering\arraybackslash}p{3.3cm}  } \toprule

\multicolumn{2}{ c }{\textbf{LivDet 2017}} & \multicolumn{2}{ c }{\textbf{Fingerprint Spoof Buster~\cite{chugh2019fingerprint}}} & \multicolumn{2}{ c }{\textbf{Fingerprint Spoof Buster + UMG wrapper}} \\ \midrule

\textbf{Sensor in Training} & \textbf{Sensor in Testing} & \textbf{ACE (\%)} & \textbf{TDR @ FDR = 1.0\%} & \textbf{ACE (\%)} 	& \textbf{TDR @ FDR = 1.0\%} \\ \toprule

Green Bit 		& Orcanthus		& $49.43$ 	& $0.00$ & $\textbf{66.05}$	& $\textbf{21.52}$ \\ \midrule
Green Bit 		& Digital Persona 	& $89.37$ 	& $57.48$ & $\textbf{94.81}$	& $\textbf{72.91}$ \\ \midrule

Orcanthus 	& GreenBit 		& $69.93$ 	& $8.00$ 	& $\textbf{81.75}$ 	& $\textbf{30.91}$ 	\\ \midrule
Orcanthus 	& Digital Persona 	& $57.99$		& $4.97$ 	& $\textbf{76.36}$ 	& $\textbf{28.46}$ 	\\ \midrule

Digital Persona & GreenBit 		& $89.54$		& $57.06$	& $\textbf{96.35}$	& $\textbf{85.21}$ \\ \midrule
Digital Persona & Orcanthus 		& $49.32$		& $0.00$  	& $\textbf{68.44}$	& $\textbf{20.38}$	\\ \toprule

\multicolumn{2}{ c }{\textbf{Mean $\pm$ s.d.} } &  $67.60 \pm 18.53$ & $21.25 \pm 28.07$ & $\textbf{80.63} \pm \textbf{12.88}$  & $\textbf{43.23} \pm \textbf{28.31}$   \\
\bottomrule
\end{tabular}
}
\end{table*}

\subsection{Datasets}
The following datasets have been utilized in this study:

\subsubsection{MSU Fingerprint Presentation Attack Database (FPAD) v2.0}
A database of $5,743$ live and $4,912$ spoof images captured on CrossMatch Guardian 200\footnote{https://www.crossmatch.com/wp-content/uploads/2017/05/20160726-DS-En-Guardian-200.pdf}, one of the most popular slap readers. The database is constructed by combining the publicly available~\cite{chugh2018fingerprint} MSU Fingerprint Presentation Attack Dataset v1.0 (MSU-FPAD v1.0) and Precise Biometrics Spoof-Kit Dataset (PBSKD). Tables~\ref{tab:database} and~\ref{tab:msuresults} presents the details of this database including the sensors used, $12$ spoof materials, total number of fingerprint impressions, and the number of minutiae-based local patches for each material type. Fig.~\ref{fig:spoofs} presents sample fingerprint spoof images fabricated using the 12 materials.

\subsubsection{LivDet Datasets}
LivDet 2017~\cite{mura2018livdet} dataset is one of the most recent\footnote{The testing set of LivDet 2019 database has not yet been made public.} publicly-available LivDet datasets, containing over $17,500$ fingerprint images. These images are acquired using three different fingerprint readers, namely Green Bit, Orcanthus, and Digital Persona. Unlike other LivDet datasets, spoof fingerprint images included in the test set are fabricated using new materials (Wood Glue, Ecoflex, and Body Double), that are not used in the training set (Wood Glue, Ecoflex, and Body Double). Table~\ref{tab:database} presents a summary of the LivDet 2017 dataset.

\subsection{Minutiae Detection and Patch Extraction}
\label{sec:minu}
The proposed UMG wrapper is trained on local patches of size $96 \times 96$ centered and aligned using minutiae points. We extract fingerprint minutiae using the algorithm proposed in~\cite{cao2019end}. For a given fingerprint image $I$ with $k$ detected minutiae points, $M = \{m_1, m_2, \dots, m_k\}$, where $m_i = \{x_i,y_i,\theta_i\}$, \textit{i.e.} the minutiae $m_i$ is defined in terms of spatial coordinates ($x_i$, $y_i$) and orientation ($\theta_i$), a corresponding set of $k$ local patches $L = \{l_1, l_2, \dots, l_k\}$, each of size $[96 \times 96]$, centered and aligned using minutiae location ($x_i, y_i$) and orientation ($\theta_i$), are extracted as proposed in~\cite{chugh2018fingerprint}.

\subsection{Implementation Details}
The encoder of the UMG wrapper is the first four convolutional layers ($conv1\_1$, $conv2\_1$, $conv3\_1$, and $conv4\_1$) of a VGG-19 network~\cite{simonyan2014very} as discussed in section~\ref{sec:umg}. We use weights pre-trained on ImageNet~\cite{russakovsky2015imagenet} database which are frozen during training of the UMG wrapper. The decoder mirrors the encoder with pooling layers replaced with nearest up-sampling layers, and without use of any normalization layers as suggested in~\cite{huang2017arbitrary}. Both encoder and decoder utilize reflection padding to avoid border artifacts. The discriminator for computing the adversarial loss is similar to the one used in~\cite{radford2015unsupervised}. The weights for style loss and content loss are set to $\lambda_s = 0.002$ and $\lambda_c = 0.001$. We use the Adam optimizer~\cite{kingma2014adam} with a batch size of $8$ and a learning rate of ($1e-4$) for both generator (decoder) and discriminator objective functions. The input local patches are resized from $96 \times 96$ to $256 \times 256$ as required by the pre-trained encoder based on VGG-19 network. All experiments are performed in the TensorFlow framework. 

For the spoof detector, we train a MobileNet-V1~\cite{howard2017mobilenets} classifier from scratch similar to ~\cite{chugh2018fingerprint} using the augmented dataset. The last layer of the architecture, a 1000-unit softmax layer (originally designed to predict the $1,000$ classes of ImageNet dataset), was replaced with a 2-unit softmax layer for the two-class problem, i.e. live vs. spoof. The optimizer used to train the network is RMSProp with asynchronous gradient descent and a batch size of 100.

\subsection{Experimental Protocol}
The fingerprint spoof generalization performance against unknown materials is evaluated by adopting a leave-one-out protocol~\cite{chugh2019fingerprint}. In the case of MSU FPAD v2.0 dataset, one out of the twelve known spoof materials is left-out and the remaining eleven materials are used to train the proposed UMG wrapper. The real spoof data (of eleven known materials) is augmented with the synthesized spoof data generated using the trained UMG wrapper, which is then used to train the fingerprint spoof detector \textit{i.e.} Fingerprint Spoof Buster~\cite{chugh2018fingerprint}. This requires training a total of twelve different UMG wrappers and spoof detection models each time leaving out one of the twelve different spoof materials. The $5,743$ live images in MSUFPAD v2.0 are partitioned into training and testing such that there are $1,000$ randomly selected live images in testing set and the remaining $4,743$ images in training such that there is no subject overlap between training and testing data splits. The real live data is also augmented with synthesized live data generated using another UMG wrapper trained on real live data.

In the case of LivDet 2017 dataset, the spoof materials available in the test set (Gelatin, Latex, and Liquid Ecoflex) are deemed as ``unknown" materials because these are different from the materials included in the training set (Wood Glue, Ecoflex, and Body Double). To evaluate the generalization performance, we evaluate the performance of Fingerprint Spoof Buster with and without using the UMG wrapper and compare with the state-of-the-art published results. As the LivDet 2017 dataset contains fingerprint images from three different readers, we train two UMG wrappers per sensor, one for each of the live and the spoof training datasets.

\begin{figure}[t]
\centering
\includegraphics[trim=0cm 0.7cm 0cm 0cm, width=\linewidth]{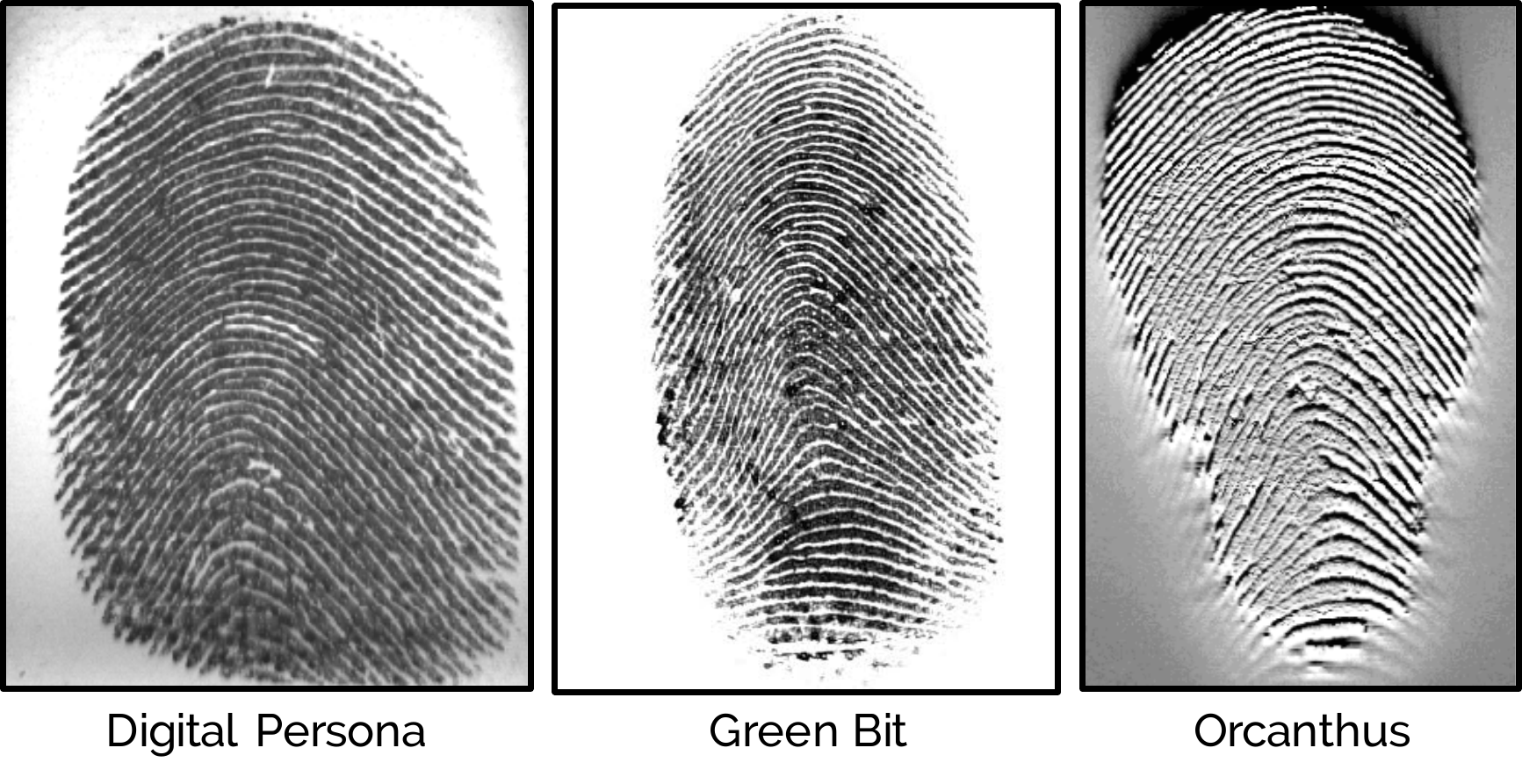}
\caption{Example fingerprint images from LivDet 2017 database captured using three different fingerprint readers, namely Digital Persona, Green Bit, and Orcanthus. The unique characteristics of fingerprints from Orcanthus reader explain the performance drop in cross-sensor scenario when Orcanthus is used as either the source or the target sensor.}
\label{fig:livdet2017}
\end{figure}

\subsection{Cross-Material Fingerprint Spoof Generalization}
Table~\ref{tab:msuresults} presents the generalization performance of the proposed approach on the MSU FPAD v2.0 dataset. The mean generalization performance of the spoof detector against unknown spoof materials improves from TDR of $75.24\%$ to TDR of $91.78\%$ @ FDR = $0.2\%$, resulting in a $67\%$ decrease in the error rate, when the spoof detector is trained in conjunction  with the proposed UMG wrapper. Table~\ref{tab:livdetresults} presents a performance comparison of the proposed approach and the state-of-the-art approach when tested on the publicly available LivDet 2017 dataset. The proposed UMG wrapper improves the state-of-the-art~\cite{chugh2018fingerprint} average cross-material spoof detection performance from 95.44\% to 95.88\%. However, a much higher performance gain is observed, from TDR of 73.32\% to 80.74\%, at a strict operating point of FDR = 1\%.

\begin{figure}[t]
\centering
\includegraphics[trim=0cm 0.5cm 0cm 0cm, width=\linewidth]{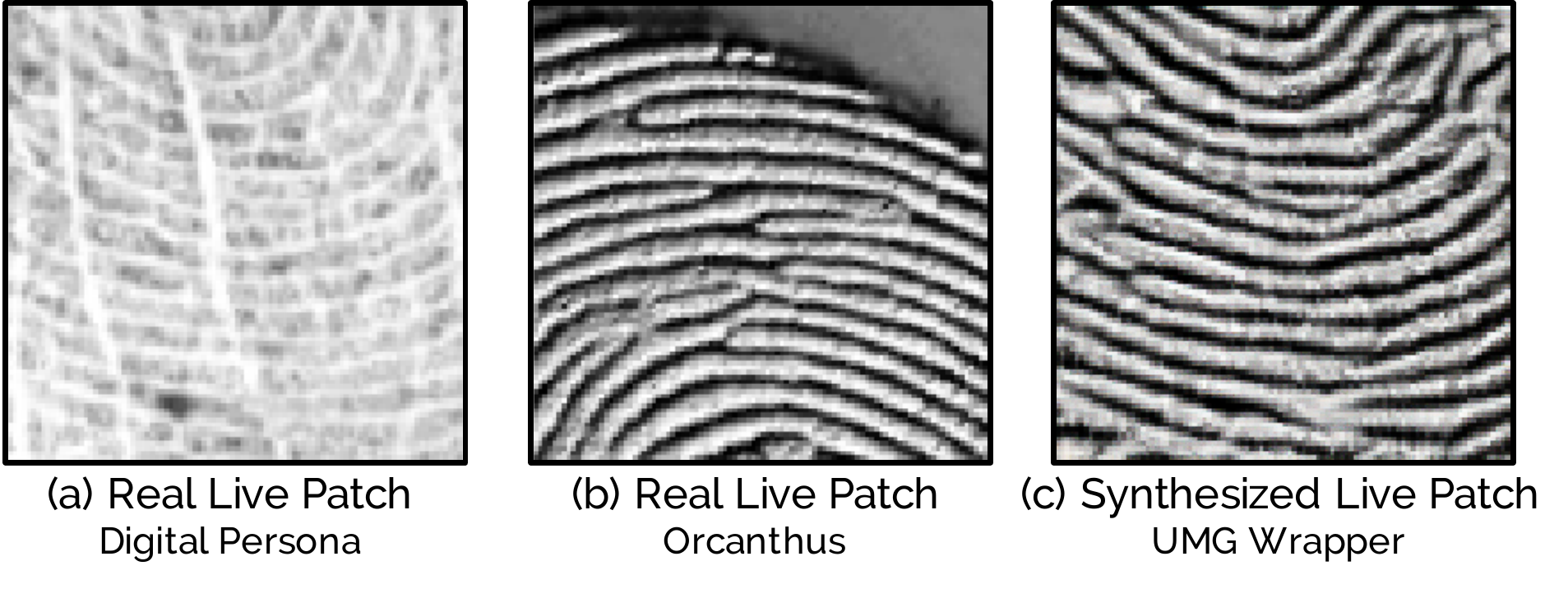}
\caption{UMG wrapper used to transfer style from (b) a real live patch from Orcanthus reader, to (a) a real live patch from Digital Persona, to generate (c) a synthesized patch.}
\label{fig:ocst}
\end{figure}

\subsection{Cross-Sensor Fingerprint Spoof Generalization}

To improve the cross-sensor performance, we employ the proposed UMG wrapper to synthetically generate large-scale live and spoof datasets to train a spoof detector for the target sensor. Given a real fingerprint database, $D^A_{real}$, collected on a source fingerprint sensor, $F^A$, containing real live, $L^A_{real}$, and real spoof $S^A_{real}$ datasets, s.t. $D^A_{real} = \{L^A_{real} \cup S^A_{real}\}$, the proposed UMG wrapper is used to generate $50,000$ synthetic live patches, $L^B_{synth}$, and $50,000$ synthetic spoof patches, $S^B_{synth}$, for a target sensor, $F_B$. The UMG wrapper is trained only on the live images collected on $S_B$, and used for style transfer on $L^A_{real}$ and $S^A_{real}$ to generate $L^B_{synth}$, and $S^B_{synth}$, respectively. We evaluate the cross-sensor generalization performance using LivDet 2017 dataset where the UMG wrapper trained on source sensor, say Green Bit, is used to generate synthetic data for a target sensor, say Orcanthus, using only a small set of $100$ live fingerprint images from the target sensor\footnote{An average of $\sim3100$ local patches are extracted from $100$ live fingerprint images in LivDet 2017 experiments.}. The spoof detector is trained from scratch only on the synthetic dataset created for the target sensor using UMG wrapper and tested on the real test set of the target sensor. Table~\ref{tab:crosssensor} presents the cross-sensor fingerprint spoof generalization performance of the spoof detector in terms of average classification accuracy and TDR (\%) @ FDR = $1\%$. We note that the proposed UMG wrapper improves the average cross-sensor spoof detection performance from 67.60\% to 80.63\%. Figure~\ref{fig:livdet2017} presents example fingerprint images captured using the three sensors in LivDet 2017. The unique characteristics of fingerprints from Orcanthus reader explain the performance drop in cross-sensor scenario when it is used as either the source or the target sensor.

\begin{figure}[t]
\centering
\includegraphics[trim=0cm 0.6cm 0cm 0cm, width=\linewidth]{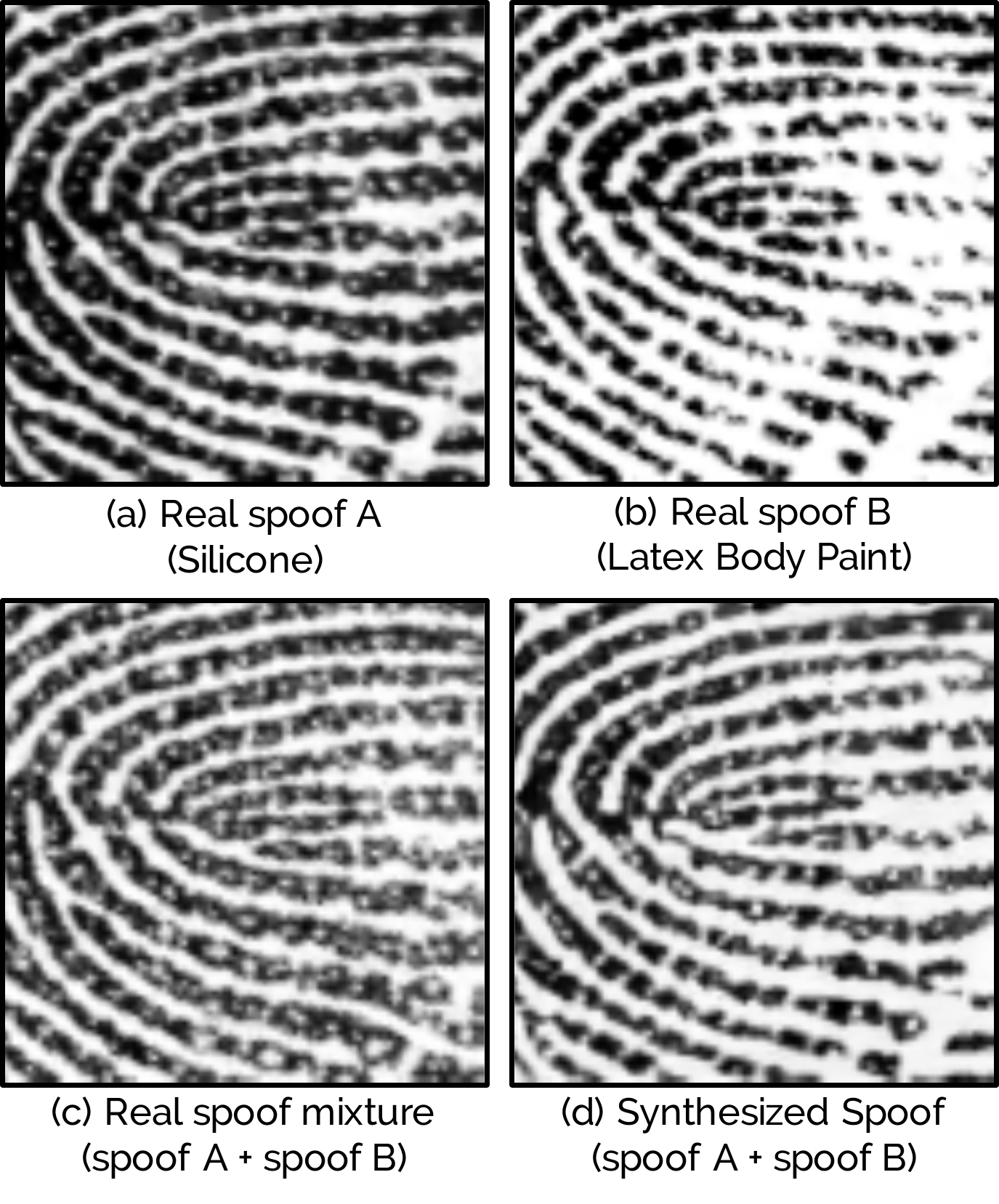}
\caption{Fingerprint patches fabricated with real spoofs (a) silicone, (b) latex body paint, (c) their mixture (in 1:1 ratio), and (d) synthesized using UMG wrapper with style transfer between silicone and latex body paint.}
\label{fig:spoof_fab}
\end{figure}

\subsection{Computational Requirements}
The proposed approach includes an offline stage of training the UMG wrapper and synthesis of fingerprints for augmenting the training dataset. Therefore, once the spoof detector is trained on the augmented data, the proposed approach has no impact on the computational requirements in the online spoof detection test stage. The proposed UMG wrapper takes under $2$ hours to train, and around $1$ hour to generate 100,000 local fingerprint patches on a Nvidia GTX 1080Ti GPU.

\section{Fabricating Unknown Spoofs}
To explore the role of cross-material style transfer in improving generalization performance, we fabricate physical spoof specimens using two spoof materials, namely silicone and latex body paint, and their mixture in a 1:1 ratio by volume\footnote{Not all spoof materials can be physically combined and may result in mixtures with poor physical properties for them to be used to fabricate any good quality spoof artifacts.}. We fabricate a total of $24$ physical specimens, including $8$ specimens for each of the two materials, and $8$ specimens using their mixture. A total of $72$ spoof fingerprints, $3$ impressions/specimen, are captured using a CrossMatch Guardian 200 fingerprint reader. Fingerprint Spoof Buster, trained on twelve known spoof materials including silicone and latex body paint, achieves TDR of 100\% @ FDR = 0.2\% on the two known spoof materials, and TDR of 83.33 @ FDR = 0.2\% against the mixture. We utilize the testing dataset of $1,000$ live fingerprint images from MSU FPAD v2.0 for these experiments.

\begin{figure}[t]
\centering
\includegraphics[width=\linewidth]{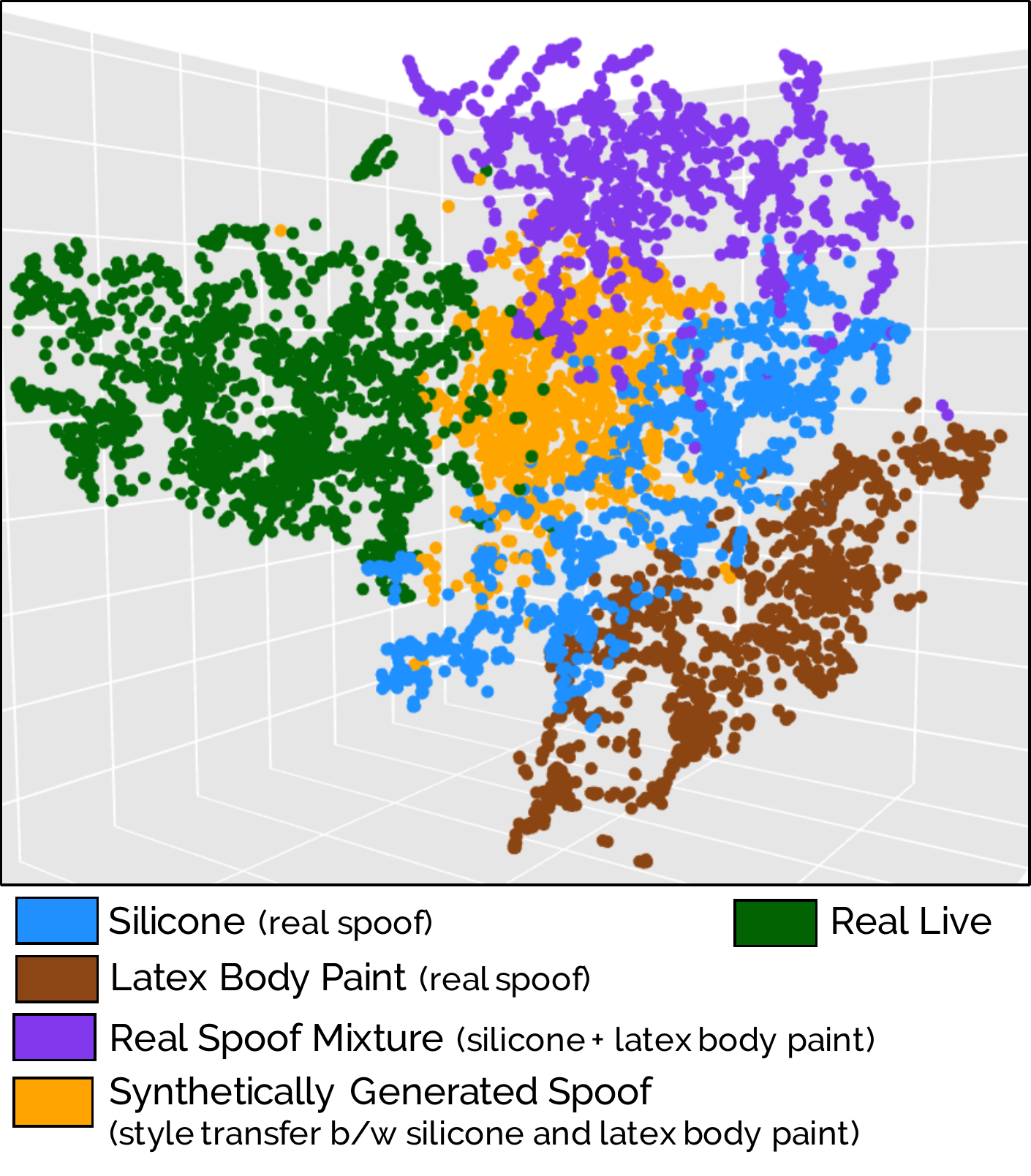}
\caption{3D t-SNE visualization of feature embeddings of real live fingerprints, spoof fingerprints fabricated using silicone, latex body paint, and their mixture (1:1 ratio), and synthesized spoof fingerprints using style-transfer between silicone and latex body paint spoof fingerprints. The 3D embeddings are available at http://tarangchugh.me/posts/umg/index.html}
\label{fig:cross_mat_tsne}
\end{figure}

We utilize the proposed UMG wrapper to generate a dataset of $5,000$ synthesized spoof patches\footnote{Around 1,100 minutiae-based local patches are extracted from 24 fingerprint images corresponding to each material.} using cross-material style transfer between spoof fingerprints of silicone and latex body paint. Fingerprint Spoof Buster, fine-tuned using the synthesized dataset, improves the TDR from 83.33\% to 95.83\% @ FDR = 0.2\% when tested on the silicone and latex body paint mixture, highlighting the role of the style-transferred synthesized data in improving generalization performance. Figure~\ref{fig:spoof_fab} presents sample fingerprint patches of the two spoof materials, silicone and latex body paint, their physical mixture, and synthesized using style-transfer. Figure~\ref{fig:cross_mat_tsne} presents the 3D t-SNE visualization of feature embeddings of live fingerprints (green), two materials, silicone (blue) and latex body paint (brown), their mixture (purple), and synthetically generated images (orange). Although the mixture embeddings are not exactly in between the embeddings for the two known materials, possibly due to low-dimensional t-SNE representation, they are close to the embeddings of the synthetically generated spoof images. This explains the improvement in performance against mixture when synthesized spoofs are used in training. Therefore, the proposed UMG wrapper is able to generate spoof images that are potentially similar to the unknown spoofs.

\section{Conclusions}
Automatic fingerprint spoof detection is critical for secure operation of a fingerprint recognition system. Introduction of new spoof materials and fabrication techniques poses a continuous threat and requires design of robust and generalizable spoof detectors. To address that, we propose a style-transfer based wrapper, Universal Material Generator (UMG), to improve the generalization performance of any spoof detector against novel spoof fabrication materials that are unknown to the system during training. The proposed approach is shown to improve the average generalization performance of a state-of-the-art spoof detector from TDR of 75.24\% to 91.78\% @ FDR = 0.2\% when evaluated on a large-scale dataset of 5,743 live and 4,912 spoof images fabricated using 12 materials. It is also shown to improve the average cross-sensor performance from 67.60\% to 80.63\% when tested on LivDet 2017 dataset, alleviating the time and resources required to generate large-scale spoof datasets for every new sensor. We have also fabricated physical spoof specimens using a mixture of known spoof materials to explore the role of cross-material style-transfer in improving generalization performance.

\section*{Acknowledgment}
This research is based upon work supported in part by the Office of the Director of National Intelligence (ODNI), Intelligence Advanced Research Projects Activity (IARPA), via IARPA R\&D Contract No. $2017-17020200004$. The views and conclusions contained herein are those of the authors and should not be interpreted as necessarily representing the official policies, either expressed or implied, of ODNI, IARPA, or the U.S. Government. The U.S. Government is authorized to reproduce and distribute reprints for governmental purposes notwithstanding any copyright annotation therein.

\ifCLASSOPTIONcaptionsoff
  \newpage
\fi




\bibliographystyle{IEEEtran} 
\bibliography{generalization.bib}

\begin{IEEEbiography}[{\includegraphics[width=1in,height=1.25in,clip,keepaspectratio]{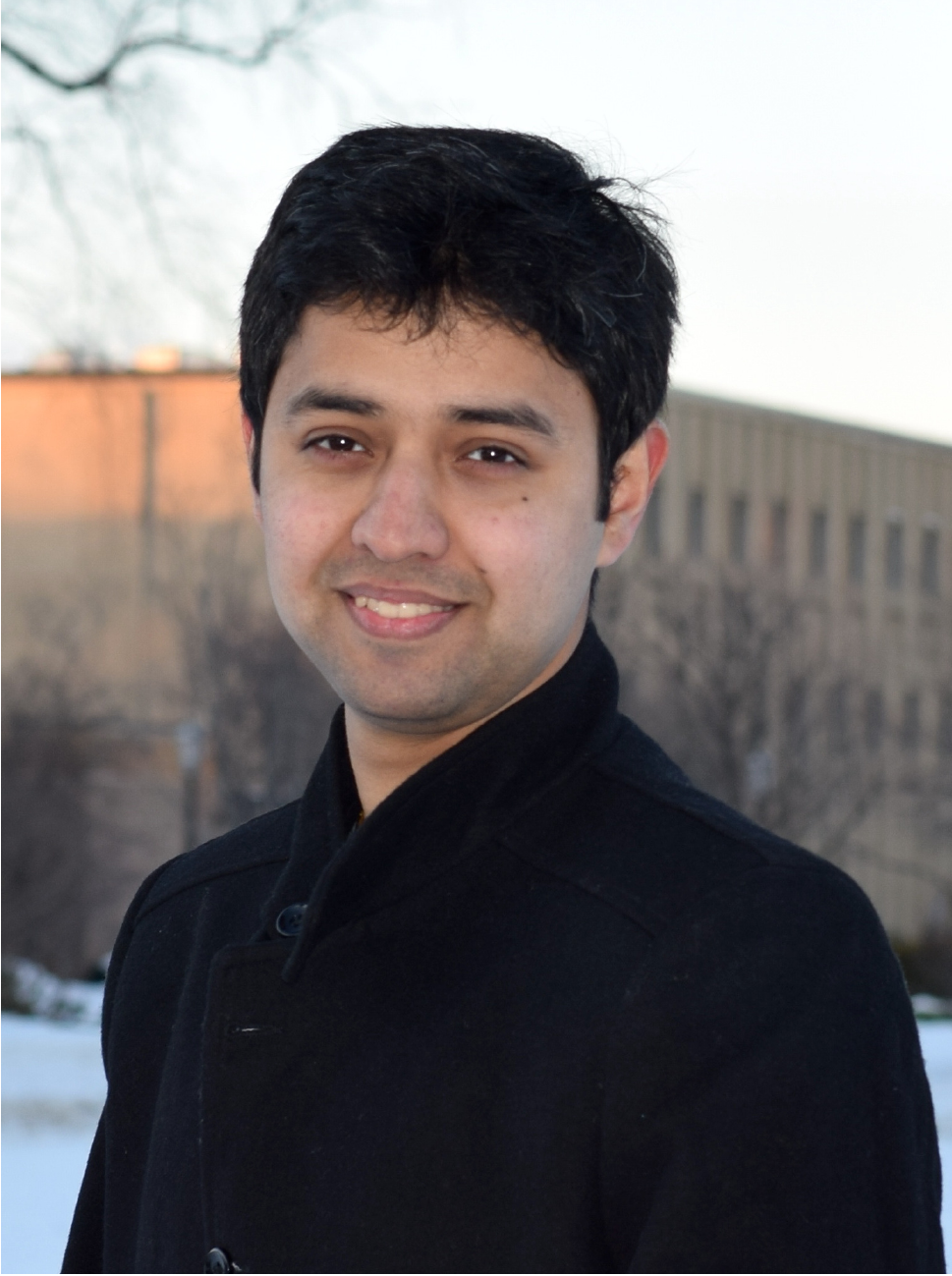}}]{Tarang Chugh}
received the B. Tech. (Hons.) degree in Computer Science and Engineering from the Indraprastha Institute of Information Technology, Delhi (IIIT-D) in 2013. He was affiliated with IBM Research Lab, New Delhi, India as a research engineer during 2013-2015. He is currently a doctoral student in the Department of Computer Science and Engineering at Michigan State University. His research interests include biometrics, pattern recognition, and machine learning.
\end{IEEEbiography}

\begin{IEEEbiography}[{\includegraphics[width=1in,height=1.25in,clip,keepaspectratio]{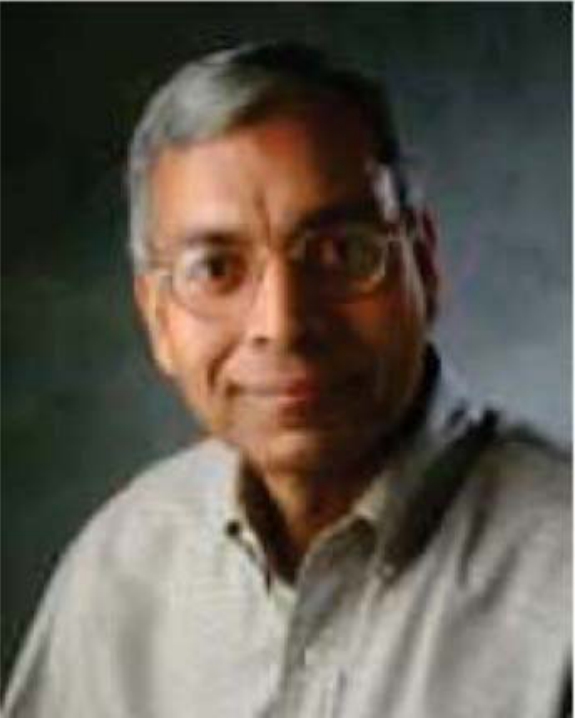}}]{Anil K. Jain}
is a University distinguished professor in the Department of Computer Science and Engineering at Michigan State University. His research interests include pattern recognition and biometric authentication. He served as the editor-in-chief of the IEEE Transactions on Pattern Analysis and Machine Intelligence and was a member of the United States Defense Science Board. He has received Fulbright, Guggenheim, Alexander von Humboldt, and IAPR King Sun Fu awards. He is a member of the National Academy of Engineering and foreign fellow of the Indian National Academy of Engineering and Chinese Academy of Sciences.
\end{IEEEbiography}

\end{document}